%% file: ms.tex
\pdfoutput=1
\documentclass[sigconf,natbib]{acmart}
\renewcommand\footnotetextcopyrightpermission[1]{}

\copyrightyear{2018}
\acmYear{2018}
\setcopyright{acmlicensed}
\acmConference[]{}{}{} 
\acmDOI{}
\acmISBN{}
\acmPrice{}

\usepackage[subtle, wordspacing=normal, tracking=normal, bibnotes, charwidths=normal, leading, indent, lists, paragraphs=normal, mathspacing=normal, bibliography=normal]{savetrees}
\usepackage{booktabs,tabularx}
\usepackage[skip=0pt]{caption}
\usepackage[skip=0pt]{subcaption}
\usepackage{caption}
\usepackage{graphicx}
\usepackage{rotating}
\usepackage{graphicx}
\usepackage{bm}
\usepackage{graphics}
\usepackage{pgfplots}
\usepackage{float}
\usepackage{setspace}
\usepackage{paralist}
\usepackage{enumitem}
\usepackage{algorithm, algpseudocode}
\usepackage{xcolor,colortbl}
\usepackage{setspace}

\title{Faithfully Explaining Rankings in a News Recommender System}

\author[M. ter Hoeve]{Maartje ter Hoeve}
\authornote{Research performed while intern at Blendle.}
\orcid{}
\affiliation{%
	\institution{University of Amsterdam}
	\city{Amsterdam}
	\country{The Netherlands}
}
\email{m.a.terhoeve@uva.nl}

\author[A. Schuth]{Anne Schuth}
\authornote{Now at De Persgroep, Amsterdam.}
\orcid{0000-0002-5841-2134}
\affiliation{%
	\institution{Blendle Research}
	\city{Utrecht}
	\country{The Netherlands}
}
\email{anne.schuth@gmail.com}

\author[D. Odijk]{Daan Odijk}
\authornote{Now at RTL.}
\orcid{}
\affiliation{%
	\institution{Blendle Research}
	\city{Utrecht}
	\country{The Netherlands}
}
\email{daan@blendle.com}

\author[M. de Rijke]{Maarten de Rijke}
\orcid{0000-0002-1086-0202}
\affiliation{%
	\institution{University of Amsterdam}
	\city{Amsterdam}
	\country{The Netherlands}
}
\email{derijke@uva.nl}

\begin{document}
	
	\begin{abstract}
	There is an increasing demand for algorithms to explain their outcomes. 
	So far, there is no method that explains the rankings produced by a ranking algorithm. 
	To address this gap we propose LISTEN, a \textbf{LIST}wise \textbf{E}xplai\textbf{N}er, to explain rankings produced by a ranking algorithm. 
	To efficiently use LISTEN in production, we train a neural network to learn the underlying explanation space created by LISTEN; we call this model Q-LISTEN. 
	We show that LISTEN produces faithful explanations and that Q-LISTEN is able to learn these explanations. 
	Moreover, we show that LISTEN is safe to use in a real world environment: users of a news recommendation system do not behave significantly differently when they are exposed to explanations generated by LISTEN instead of manually generated explanations.
	\end{abstract}
	
%

        
	
	\maketitle
	
	\input{01-introduction.tex}
	\input{02-related-work.tex}
	\input{03-method.tex}
	\input{04-experimental-setup.tex}
	\input{05-results.tex}
	\input{06-discussion-and-conclusion.tex}

    \begin{spacing}{1}
\small\noindent\scriptsize
\\
	\textbf{Acknowledgements}
This research was supported by
Ahold Delhaize,
Amsterdam Data Science,
the Bloomberg Research Grant program,
the China Scholarship Council,
the Criteo Faculty Research Award program,
Elsevier,
the European Community's Seventh Framework Programme (FP7/2007-2013) under
grant agreement nr 312827 (VOX-Pol),
the Google Faculty Research Awards program,
the Microsoft Research Ph.D.\ program,
the Netherlands Institute for Sound and Vision,
the Netherlands Organisation for Scientific Research (NWO)
under pro\-ject nrs
CI-14-25, 
652.\-002.\-001, 
612.\-001.\-551, 
652.\-001.\-003,
and
Yandex.
All content represents the opinion of the authors, which is not necessarily shared or endorsed by their respective employers and/or sponsors.
\end{spacing}

\bibliographystyle{ACM-Reference-Format}
\clearpage
\bibliography{ms}
	
\end{document}

%% file: 01-introduction.tex

\section{Introduction}
\label{sec:introduction}
There is an increasing demand for data-driven methods to be explainable. This has especially become relevant these days, since on the 14th of April 2017, the General Data Protection Regulation (GDPR) was approved by the EU parliament and it will be enforced on the 25th of May, 2018. Amongst others the GDPR states that we need to be able to explain algorithmic decisions.
Explainability of machine learning algorithms has received considerable attention from the research community \citep[e.g.,][]{bilgic2005explaining, herlocker2000explaining, tintarev2007survey, musto2016explod, vig2009tagsplanations}. 
In the context of information retrieval, research by \citet{ter2017news} shows that users clearly state that they would like to receive explanations for their personalized news selection, which is presented to them as a ranked list. 
Despite this, the explainability of ranking algorithms has never been fully addressed. 

Explaining a ranking is the challenge that we address in this paper. Previous research has focussed on explaining single data points. E.g., one could focus on the explanation of a single recommendation or a single classification~\citep[e.g.,][]{ribeiro2016should, hechtlinger2016interpretation, ross2017right}. However, a ranking can only be explained by looking at all items in the ranking --- the position of an item in a ranking is dependent on the other items that also occur in this ranking. 

Importantly, any method to explain a ranking should faithfully explain the outcome of the ranking algorithm. 
By \textit{faithful} we mean that we want our explainer to solely base its explanations on the underlying structure of the algorithm. 
Naturally, the best explanation of an algorithm is the underlying structure of the algorithm itself. However, even for experts in the field, these explanations may be uninterpretable. 
Therefore, we also want our explanations to be \textit{interpretable}. 
\citet{doshi2017towards} define \textit{interpretability} as: ``\textit{the ability to explain or to present in understandable terms to a human}.'' 
In this spirit we aim to find the most important causes of an event that can be mapped directly to a human understandable message.\footnote{While automatically generating a natural language statement as an explanation of a ranking is an interesting research direction, it is not part of the focus of this paper.}
We aim to be able to explain the rankings produced by any type of ranking algorithm and as such we also want any proposed explainer to be \textit{model-agnostic}. 

In this paper, we introduce a faithful approach to explain ranking algorithms: LISTEN --- a \textbf{LIST}wise \textbf{E}xplai\textbf{N}er. 
The design of LISTEN is based on the intuition that we can find the importance of ranking features by perturbing their values and by measuring to what degree the ranking changes due to that.
Subsequently, we design and train a neural network, Q-LISTEN, that learns explanations generated by LISTEN and is sufficiently efficient to run in a production environment.
In other words, we contribute an explanation pipeline for rankings that can run in real-time and that can therefore be used in real-life applications. 

We address the following research questions:

\begin{description}[nosep]
	\item[RQ1] Do LISTEN and Q-LISTEN produce faithful explanations of rankings?
	\item[RQ2] Does the type of explanation affect the users' behavior? 
\end{description}

\noindent%
As to RQ2, we are keen to find out whether the reading behavior of users who are provided with faithful and model-agnostic explanations for a personalized ranked selection of news articles differs from the reading behavior of users who are provided with heuristic explanations for their personalized ranked selection of news articles. Our goal is to provide users with faithful explanations of the occurrence of items in their rankings. It is \emph{not} our goal to affect users' reading behavior by providing them with explanations.

In the remainder of this paper, we first describe the difference between explaining a ranking and explaining a single item in a ranking in more detail. 
We also present the problem setting in which we conducted this research. 
In Section~\ref{sec:related_work}, we present the relevant related work and in Section~\ref{sec:design_listen} we present our design of LISTEN and Q-LISTEN. 
Section~\ref{sec:experimental_setup} gives our experimental setup. 
In Section~\ref{sec:results}, we present our results and answer our research questions. 
We end with a discussion and conclusion in Section~\ref{sec:disc_and_con}. 

\subsection{Approaches to explaining rankings}
To motivate our work and get an intuition for our approach, imagine a ranking algorithm that uses a simple linear ranking scoring function to compute the relevance of particular items. The ranking function is given by 
\begin{equation}
\label{eq:simple_scoring1}
score(x_0, x_1, x_2) = 0.2 x_0 + 0.3 x_1 + 0.5 x_2,
\end{equation}
where $x_0$, $x_1$ and $x_2$  are features. 
In a real application these could be features that describe characteristics of the item, the user, general features such as the current season or time, etc. 
In this example, the feature $x_0$ and $x_1$ can take on values in the range $[0, 1]$ and $x_2$ can take on values in the range $[0.6, 1]$. 
Assume that we have a ranking with three items described by the feature value matrix 
\[
	\begin{array}{l l l l | l }
	& x_0	& x_1	& x_2 & score \\
	d_0 & 1   & 1   & 1   & 1    \\
	d_1 & 0.5 & 0.5 & 1   & 0.75 \\
	d_2 & 1   & 0   & 0.7 & 0.55
	\end{array} 
\]
where the last column is the score computed by Eq.~\ref{eq:simple_scoring1} and $d$ stands for \textit{document}.
Our task is to explain this ranking. 
There are at least two possible approaches. 
We could focus on a single document and its corresponding score and mark the feature that contributed most to the score as the most important feature and, hence, give this feature as explanation for why this document is selected for this ranking. 
This is a \textit{pointwise explanation}, because it only takes one item, i.e., one point, in the ranking into account when explaining the occurrence of that item in the ranking. 
One important shortcoming of this approach is that it does not explain the \textit{rank} of a particular item --- it just explains its score. In order to explain the rank of an item, one needs to take all other items in the ranking into account as well. 
This is the \textit{listwise} approach, because it considers the entire list of items for its explanations. 
Below we give an example to show the difference between the two approaches: the \textit{pointwise} approach on the one hand and the \textit{listwise} approach on the other hand.

We use the feature value matrix that we introduced above and we want to find the most important feature for the first item in the ranking, $d_0$. 
A pointwise approach would mark feature $x_2$ as most important, as this feature value, together with its corresponding weight, contributes most to the score of the first document. 
In contrast, a listwise method would mark feature $x_1$ as most important, because feature $x_1$ is able to change the ranking, whereas the feature $x_2$ is not. If we change the feature value of feature $x_2$ to $0.6$, the lowest possible value, the score of $d_0$ becomes $0.8$, which still places $d_0$ on top of the list. 
But if we change the value of feature $x_1$ to its lowest possible value, namely $0$, the score of $d_0$ becomes $0.7$ which places $d_0$ below $d_1$ and hence changes the ranking. 
This is the behavior that we want to capture in our explanations.

We can construct a similar example if we look at $d_2$. Again, the pointwise explanation would mark $x_2$ as the most important feature, as this feature value and its weight make the score go up most. A listwise explanation would mark feature $x_1$ as the most important feature, something a pointwise explanation would not do, as $0.3 \cdot 0 = 0$. However, a listwise explanation would find that feature $x_2$ is not able to change the ranking. Changing it to the largest possible value, $1$, would change the score to $0.7$, and changing $x_2$ to its lowest possible value would make the score be $0.45$; both changes leave the ranking as it is. But changing $x_1$ to $1$ would give $d_2$ the second position in the ranking, above $d_1$ as then the score would become $0.85$. 

These two examples show that a pointwise explanation method does not capture the behavior that we want to explain. 
Alternatively, a \textit{pairwise} explanation method, where we would only compare pairs of items in the ranking, would not suffice either; for similar reasons as in the pointwise case, this would not allow us to capture the behavior that we want to explain. In contrast, listwise explanations do capture the right behavior.
Another observation that motivates us to design listwise explanations is that many-state-of-the-art ranking algorithms are optimized to learn an entire ranking, instead of individual scores of items in a ranking. 
Therefore, a listwise explanation style is the only way to provide faithful explanations for these types of ranking functions. Here we only considered two examples, but similar reasoning holds for more complex ones.

We aim to develop a faithful listwise explanation method and compare this to a heuristic pointwise explanation method baseline. In the context of ranking algorithms, we define a \textit{pointwise explanation} to be an explanation that only takes the score of an individual item into account. We define a \textit{listwise explanation} to be an explanation that takes the entire ranking into account. What a listwise explanation could look like in practice is non-trivial. We address this question in Section~\ref{sec:design_listen}. We develop and test our approach to construct listwise explanations on a production news recommender system. Below we briefly describe our problem setting. 

\subsection{Problem setting}
\label{subsection:problem-setting}
We conduct this study in the setting of Blendle. Every day, Blendle users receive a personalized selection of news articles from a wide variety of newspapers. 
These articles are selected based on a number of features that capture users' reading behavior and topical interests. These features are summarized in Table~\ref{tab:features_number_mapping}. 
Blendle has performed a feature analysis to make sure these features are uncorrelated. To the best of our knowledge, the approach Blendle takes is representative for many personalized recommender systems that run in production.
On top of this, Blendle users also receive a number of \textit{must reads} every day; these articles are selected by the editorial staff and are the same for everyone. 
This is one of the ways to help prevent users from ending up in their own filter bubble. 
The editorial staff manually writes a small summary, or recommendation, for each of the selected articles that users can read before they decide to open the article. 
Blendle allows users to purchase a single news article instead of having to buy an entire newspaper (using micropayments) or to prepay via a subscription for their personal selection. 
Users have the possibility to receive a refund for an article if they are not satisfied with it. 

\begin{table}
	\centering
	\caption{Features used by the production news recommender at the time of writing, numbered.}
	\label{table-att}
	\begin{tabular}{ll}
		\toprule
		\textbf{Feature}       & \textbf{Feature description}\\ 
		\midrule
		$f_0$ & item rating score      \\
		$f_1$ & item pick probability   \\ 
		$f_2$ & item number of images   \\
		$f_3$ & item topic followed by user \\ 
		$f_4$ & item newspaper followed by user \\ 	           
		$f_5$ & user purchased topic score \\
		$f_6$ & user purchased newspaper score \\
		$f_7$ & user item negative topic feedback \\
		$f_8$ & user item negative newspaper feedback \\
		\bottomrule                    
	\end{tabular}
	\label{tab:features_number_mapping}
    \vspace{-10pt}
\end{table}

%% file: 02-related-work.tex

\section{Related work}
\label{sec:related_work}
The notion of \textit{explanation} and its goal has been the subject of many studies, especially in the social sciences. \citet{miller-explainable-2017} and~\citet{doshi2017towards} give an extensive overview of this research and how it can be applied to the field of artificial intelligence. 
Based on this overview, we define the goal of an explanation in this research to \textit{faithfully give the underlying cause of an event}. 
In Section~\ref{sec:introduction} we introduced the notion of \textit{faithful}. 
To define the notion more thoroughly we build on \citep{vig2009tagsplanations}, where two kinds of explanation styles are introduced: \textit{justifications} and \textit{descriptions}. 
Justifications focus on providing conceptual explanations that do not necessarily expose the underlying structure of the algorithm, whereas descriptions are meant to do exactly that. 
We aim to provide descriptions instead of justifications, as one of our main goals is to provide faithful explanations, that are solely based on the underlying structure of the algorithm. 
Descriptions can be \emph{local} or \emph{global}. 
Local descriptions only explain the underlying structure of a particular part of the model, whereas global descriptions aim to explain the entire model, thereby not allowing for simplifications of the model by only looking at a particular part of the model. 
We aim to construct global explanations, as this increases the faithfulness of the explanation.

Below, we present related work on explainability in machine learning, on feature selection and on learning to rank. 

\subsection{Explainability in machine learning}

Previously, many studies that focus on the explainability of machine learning algorithms have been conducted from a Human Computer Interaction angle~\citep[e.g.,][]{herlocker2000explaining, bilgic2005explaining, tintarev2007explaining, ter2017news}. That is, questions are asked such as ``how do users interact with the system and how can explanations help with this?'' These studies do not focus on how to construct \textit{faithful} explanations to describe the underlying decisions of the algorithm. Instead, explanations are made up to give users an idea of what the explanations could be like. Recently the focus is changing towards describing the training process~\cite[e.g.,][]{koh-2017-understanding,sharchilev-finding-2018} and towards the underlying algorithm~\citep[e.g.,][]{musto2016explod, vig2009tagsplanations, abdollahi2017accurate, hendricks2016generating}. The latter helps to increase the faithfulness of the explanations~\citep{pu2007trust}.

\citet{hechtlinger2016interpretation} and~\citet{ross2017right} use the gradients of the output probability of a model with respect to the input to define feature importance in a predictive model. The importance scores are used to interpret the behavior of the model. This is an intuitive approach, yet one important prerequisite of using this method is that the models are differentiable with respect to their inputs. However desirable, this is not a property of all models. For example, the state-of-the-art LambdaMart ranking algorithm~\citep{burges2010ranknet} lacks this property. The approaches by \citeauthor{hechtlinger2016interpretation} and~\citeauthor{ross2017right} focus on a single data point --- they are \textit{pointwise} instead of \textit{listwise}.

\citet{ribeiro2016should} introduce LIME, a method that can be used to locally explain the classifications of any classifier. 
Three important characteristics underlie the construction of LIME: an explaining model needs to be (1) ``\textit{interpretable},'' (2)~``\textit{locally faithful},'' and (3) ``\textit{model-agnostic},'' which~\citeauthor{ribeiro2016should} define as (1)~``\textit{provide qualitative understanding between the input variables and the response},'' (2)~the explanation ``\textit{must correspond to how the model behaves in the vicinity of the instance being predicted},'' and (3)~``\textit{the explanation should be able to explain any model}, respectively. \citeauthor{ribeiro2016should} provide linear models, decision trees and falling rule lists as examples of interpretable models. 
There are two important reasons why we \emph{cannot} use LIME to explain a ranking. First, LIME is designed to explain the decisions of classifiers, whereas we aim to explain a ranking function. Secondly, even if we would adapt LIME in such a way that we treat the ranking function as a classifier (for example by binning the outputs) LIME only aims to be locally faithful and therefore it will produce pointwise explanations instead of the listwise explanations that we aim for.

\subsection{Feature selection}
We can use some of the intuitions that are used in the feature selection research to solve our current problem. 
The goal of feature selection is to find a relevant subset of features for a model. 
There is a substantial amount of research on this topic~\citep[e.g.,][]{geng2007feature, lai2013fsmrank, hua2010hierarchical, laporte2014nonconvex, dash1997feature, battiti1994using}. 
Many studies aim to find the set of features that maximize the importance of the features in the set and minimize the similarity of features in the set. 
Finding the importance scores for features is related to the explainability question addressed in this paper. 
The difference is that we try to find features that are important for an item's position in the ranking, whereas feature selection techniques aim to find important features for the entire set.

\citet{battiti1994using} uses Shannon's entropy~\citep{shannon1951mathematical} to select new features for classification problems. Features that contain most information and therefore decrease the uncertainty about a classification are selected. Several studies use dimensionality reduction techniques such as PCA for feature selection~\citep{malhi2004pca, yu2003feature}. \citet{geng2007feature} design a feature selection method for ranking. They measure the importance of features by metrics such as MAP, NDCG and loss functions such as pairwise ranking errors. 
Similarity between features is measured by measuring similarity of the resulting rankings. 
\citet{hua2010hierarchical} compute feature similarity in the same fashion. 
After that, they cluster features based on their similarity scores. 
Only a single feature from each cluster is selected. We use ranking similarity as a metric to measure how features are able to change a ranking.

\subsection{Learning to rank}

Ranking is used in several domains~\citep[e.g.,][]{page1999pagerank, haveliwala2003topic, del2005ranking}, from building search engine result pages, where a user has a specific query for the search engine, to domains in which a user has a less specific query yet is expecting to see results, such as the timelines on social networks, or personalized news selection of news as in our work. 
Producing effective ranking algorithms is the aim of learning to rank.
Learning to rank approaches can be divided into \textit{pointwise} approaches, \textit{pairwise} approaches and \textit{listwise} approaches~\citep{liu2009learning}. 
Pointwise approaches compute a relevance score for every single item that is to be ranked individually. The items are then ranked in a decreasing order of scores. Pairwise approaches look for disordered pairs in a ranking, put them in the correct order, until all pairs are ranked correctly, and thus the entire ranking as well. 
Listwise approaches try to optimize the order of the entire list at once and have information retrieval measures such as NDCG~\citep{jarvelin2002:cumulated} as their optimization objective.

%% file: 03-method.tex

\section{Designing LISTEN and Q-LISTEN}
\label{sec:design_listen}
In this section, we present our design for LISTEN. As LISTEN is not fast enough to run in production, we subsequently present Q-LISTEN, a \textbf{Q}uick version of LISTEN, for which we train a neural network to learn the explanation space for us. 

\subsection{Design decisions}
There are multiple ways to explain a ranking~\cite{guidotti-2018-survey}. 
One could explain the entire list at once, in a single statement. 
However, the interpretability and usefulness of this approach is questionable. 
One could also give contrasting explanations~\citep{miller-explainable-2017}. This would lead to explanations in the form of ``\textit{item A is ranked above item B, C, D, \ldots, because item A has characteristic X that item B does not have, characteristic Y that C and D do not have and it is ranked below item \ldots\ because \ldots}.'' One could also compare rankings with other rankings, e.g., ``\textit{Ranking A is shown as opposed to ranking B, because \ldots}.'' Such contrastive explanations easily extend to a large, cluttered presentation of the argumentation. 
We present explanations that give the main cause(s) of an item's position in the ranking. 
This choice ensures that the entire ranking is taken into account, whereas at the same time explanations can be generated that are easily interpretable by users or developers of the system.

The aim of this research is to find the most important \textit{features} for an item's position in the ranking. These are considered as the explanations. As we cannot provide users with these raw features and importance scores, we construct a mapping between each feature and an explanation in natural language. 
A prerequisite for this approach is to have interpretable features. 
Many recommender systems that run in production use interpretable features. 
This does not imply that our method does not work for systems that use other types of features, as we can still output the most important features. 
Constructing mappings from features to human interpretable explanations is a task that we leave for future work.

Another decision that we make is to only report features to the user that actually increase the score of an item in the ranking. 
We do this in order to avoid explanations such as ``\textit{you see this article because you do not really like $X$},'' whereas the article is actually about topic $X$. 
Even though these could be faithful explanations, they are unintuitive for users of the system. 
This approach does not mean that we only report features with the highest values. 
E.g., intuitively one can expect that if a user follows a topic or newspaper (high feature value) this should indeed increase the overall ranking score of the item, but if a user has not given any negative feedback (low feature value) this should also increase the ranking score. 

\subsection{LISTEN -- Overview}
We present the design of LISTEN: a \textbf{LIST}wise\- \textbf{E}xplai\-\textbf{N}er that is designed to return those features that were most defining for an item's position in the ranking, keeping the design considerations presented in the previous section in mind. The following intuition will be central: \textit{if changing the value of a feature for a certain item causes the ranking to substantially change, this feature was important for this item's position in the ranking, otherwise this feature was not important}. 
In order to design an algorithm that works according to this intuition we need to define (1)~how we change feature values and (2)~how we measure ranking dissimilarity. 
As to~(1), we discretize the feature value domains (which we extract from a training data set). As changing all features in all rankings according to this discretization is computationally infeasible, we first observe the behavior of feature values on a small training dataset and then select the most influential feature values for all features. These new values we use from then onwards. This is why we split our algorithm in a training phase and an explaining phase respectively. To measure ranking dissimilarity (i.e., item (2)), we choose to use the AP ranking correlation coefficient~\citep{yilmaz2008new} (rather than Kendall's $\tau$~\citep{kendall1938new} or Spearman's rank correlation coefficient~\citep{spearman1904proof}), given by
\begin{equation}
\label{eq:tau_ap}
\tau_{AP} = \frac{2}{N-1} \sum_{i=2}^N \bigg ( \frac{C(i)}{i-1} \bigg ) - 1.
\end{equation}
The AP ranking correlation coefficient focusses on the top elements in the list. 
This is important, as we deal with many articles that are scored and ranked, yet only the top $25$ items are selected for the user. During the explaining process, we cannot limit ourselves to only this top $25$ items, as this would prevent us from measuring the difference between a new feature value that causes an item to be placed at position $25$ and one that causes it to be placed below position $25$. 
However, changes in the higher regions of the list are more relevant than changes in the lower regions. The AP ranking correlation coefficient captures this. It ranges from $-1$ to $1$, whereby $-1$ means that two rankings are completely opposite and $1$ means that two rankings are exactly equal.

Algorithm~\ref{alg:overview} gives an overview of the steps taken in LISTEN. 
Below, we describe the individual steps of the pipeline in full detail. 
We start with the training phase in Section~\ref{subsec:training_phase}, followed by the explaining phase in Section~\ref{subsec:explaining_phase} and in Section~\ref{subsec:q_listen} we present a speed-up to be able to run LISTEN in production.

\begin{algorithm}
\small
	\begin{algorithmic}[1]
		\State \textbf{Training phase} (Section~\ref{subsec:training_phase})
		\State Find the importance of individual feature values by changing them and see how these changes affect the ranking.
		\State Find points of interest.
		\State \textbf{Explaining phase} (Section~\ref{subsec:explaining_phase})
		\State Use the points of interest to find the most important features by observing which changes in feature values affect the ranking most.
		\State Return the most important features.
		\State The most important features are the explanations. Return these to the users in an understandable way.
	\end{algorithmic}
	\caption{Overview LISTEN}
	\label{alg:overview}
\end{algorithm}

\subsection{LISTEN -- Training phase}
\label{subsec:training_phase}
The first step of the training phase is to find how individual feature values can affect the ranking (see Algorithm~\ref{alg:overview}). 
We will call this the \textit{disruptiveness} of feature values. 
Algorithm~\ref{alg:disruptiveness} summarizes the part of the training step where we find the disruptiveness of feature values. 
We define values each individual feature can take on. Then we can change one feature at a time according to those values and measure how this changes the ranking. 
Based on this, we define the most disruptive feature values per feature. 
We decide to only change a single feature value at a time, as the features Blendle uses are independent by design. 
If one cannot make this assumption, or if one wants to investigate how different feature values work together, more permutations should be tried, depending on the degree correlation of the features. 
Future work should look into how we can compute this in an efficient manner.

\begin{algorithm}
\small
	\begin{algorithmic}[1]
		\Function{FindMinAndMax}{} \label{line:min_max_function}
		\For{\textbf{each} feature $\in$ all features}
		\State \Call{FindMinValue}{feature}
		\State \Call{FindMaxValue}{feature}
		\EndFor
		\EndFunction
		
		\\
		
		\Function{FindDisruptiveness}{} \label{line:disruptiveness_function}
		\For{\textbf{each} feature $\in$ all features}
		\State \Call{FindSampleRange}{feature} \label{line:sample_ranges_functioncall}
		\For{\textbf{each} sample value $\in$ sample range} \label{line:sample_value_loop}
		\For{\textbf{each} ranking $\in$ all rankings} \label{line:ranking_loop}
		\For{\textbf{each} item $\in$ ranking} \label{line:item_in_ranking_loop}
		\If{feature value item $\neq$ sample value}	\label{line:check_value}				
		\State change feature value
		\State \Call{Calculate$\tau_{AP}$}{new ranking} \label{line:calculate_tau}
		\State store $\tau_{AP}$ 
		\EndIf
		\EndFor	
		\EndFor
		\State average all $\tau_{AP}$'s \label{line:tau_avg}
		\State store average $\tau_{AP}$
		\EndFor
		\EndFor
		\EndFunction
		
		\\
		
		\Function{FindSampleRange}{feature} \label{line:sample_ranges_function}
		\If{feature $\in$ discrete features} \label{line:categorical_range}
		\State \Return range(FeatureMinValue:FeatureMaxValue, BinSize) 
		\ElsIf{feature $\in$ predefined features} \label{line:predefined_range}
		\State \Return predefined values
		\Else
		\State \Return range(FeatureMinValue:FeatureMaxValue, BinSize) \label{line:continuous_range}
		\EndIf
		\EndFunction
		
		\\ 
		
		\State \Call{FindMinMax}{} \label{line:min_max_functioncall}
		\State \Call{FindDisruptiveness}{} \label{line:find_disruptiveness_functioncall}
      
	\end{algorithmic}
	\caption{LISTEN Training phase - (1) Find the disruptiveness of feature values}
	\label{alg:disruptiveness}
\end{algorithm}

Let us explain Algorithm~\ref{alg:disruptiveness} in more detail. 
As a first step towards finding how to change feature values, we find the minimum and the maximum value for each feature in our training data (line~\ref{line:min_max_functioncall} and line~\ref{line:min_max_function}). 
Now we proceed to finding the most disruptive feature values between these minimum and maximum values (line~\ref{line:find_disruptiveness_functioncall} and line~\ref{line:disruptiveness_function}). In order to do so we have to discretize our continuous ranges, for each feature, and return feature value samples (line~\ref{line:sample_ranges_functioncall} and line~\ref{line:sample_ranges_function}). For this, we distinguish between continuous features, discrete features and features with predefined values. 
An example of the latter feature type is the score the editorial staff assigns to an article: $0.3$, $0.6$ or $0.9$. 
For discrete features (line~\ref{line:categorical_range}) we select all integers between the minimum and the maximum value, unless we exceed a certain bound. 
In that case we divide the range in larger intervals and sample a single integer per interval. 
In Algorithm~\ref{alg:disruptiveness} we represent this with the variable $BinSize$. 
I.e., if the bound is not exceeded $BinSize=1$, otherwise $BinSize > 1$. This is a hyper parameter one can choose. 
For predefined feature values we only use those that are given in the data (line~\ref{line:predefined_range}). (If there are too many, one could choose to bound this as well.) 
For continuous feature values we discretize the range between the minimum and maximum values found (line~\ref{line:continuous_range}). 
How precisely we discretize this range is a hyper parameter that we can choose and is mostly motivated by computation time. 
We set this hyper parameter to $20$.
Now, we loop through the items in the rankings of all users and change their feature values one by one, according to the feature values we have just found (lines~\ref{line:sample_value_loop}--\ref{line:item_in_ranking_loop}). We only change the feature value if the sample value differs from the current feature value, as this makes the score more sensitive (line~\ref{line:check_value}). For each of these feature values we compute the $\tau_{AP}$ (line~\ref{line:calculate_tau}), according to~\eqref{eq:tau_ap}. 
We keep these values and compute the average (line~\ref{line:tau_avg}). 
This average is called the \emph{disruptive score}. 

\subsubsection{Step 2 -- Select points of interest}
By gradually changing all feature values that we have in our data as described above, we find, for each feature, the disruptiveness per sampled feature value. 
To increase the computation speed we aim to select only those feature values with the lowest disruptive scores and discard feature values with the highest disruptive scores and we aim to have some spreading over the feature value range. 
(Recall that low $\tau_{AP}$ scores represent dissimilar rankings and thus low disruptive scores describe disruptive feature values.) 
If we had any prior knowledge about the distributions of these disruptive scores, we could fit these to the disruptive scores we found and select the minima in these distributions as the most disruptive feature values. 
However, we do not have this prior knowledge. 
The approach we take instead is given in Algorithm~\ref{alg:pois_values}. 
We call the selected feature values \emph{points of interests} (or \emph{pois} in Algorithm~\ref{alg:pois_values}). 

In Algorithm~\ref{alg:pois_values} we divide the range of disruptive scores that we have found in the previous steps in bins. 
The number of bins is a hyper parameter that we can choose. 
We use $20$ bins. 
For each feature value for each feature we look up its disruptive score (``avg $\tau_{AP}$'' in Algorithm~\ref{alg:pois_values}) that we computed in the previous step. 
We compute the bin of this disruptive score (line~\ref{line:bin}). 
As high values for $\tau_{AP}$ mean that rankings are comparable and low values for $\tau_{AP}$ mean that rankings differ, we only keep feature values that yield the lowest average $\tau_{AP}$-values in their bin (line~\ref{line:bin_check1} until line~\ref{line:bin_check2}) for the points of interest. 
This way we ensure that we select a different range of feature values, whereas the choice for these feature values is still motivated by their disruptiveness. 
If feature values can only adopt a few predefined values we choose to use all these values as interest points (line~\ref{line:keep_predefined}), up until a certain number which again is a hyper parameter to tune. 

\begin{algorithm}
\small
	\begin{algorithmic}[1]
		\State initialization
		\For{\textbf{each} feature in all features}
		\If{feature $\in$ predefined features} \label{line:keep_predefined}
		\State keep predefined feature values as pois values 
		\State continue
		\EndIf
		\For{\textbf{each} feature value $\in$ all feature values}
		\State bin = \Call{ComputeBin}{avg $\tau_{AP}$} \label{line:bin}
		\If{bin $\neq$ empty AND $\tau_{AP} \leq \tau_{AP}$ in bin} \label{line:bin_check1}
		\State put $\tau_{AP}$ and feature value in bin 
		\State delete previous $\tau_{AP}$ and feature value from bin 
		\EndIf
		\If{bin == empty}
		\State put $\tau_{AP}$ and feature value in bin
		\EndIf \label{line:bin_check2}
		\EndFor
		\EndFor
		\State \Return kept feature values as pois values
	\end{algorithmic}
	\caption{LISTEN training phase - (2) Find the points of interest}
	\label{alg:pois_values}
\end{algorithm}

\subsection{LISTEN -- Explaining phase}
\label{subsec:explaining_phase}
So far we have found the disruptiveness of feature values and we have selected the points of interest from all these scores. In this section we present the part of LISTEN where we find the most important features for each item in the ranking. This is what we call the \textit{explaining phase}. We summarize the explaining phase in Algorithm~\ref{alg:explaining phase} and explain it in more detail here.

For each new ranking that comes in, we change all of its feature values (i.e., all feature values for all items in the ranking) according to the points of interest (line~\ref{line:ranking_loop}--\ref{line:pois_loop}). Then we compute the AP ranking correlation coefficients for the new ranking that arises from changing this feature value. We compute the average AP ranking correlation coefficients for this feature (line~\ref{line:value_check}--\ref{line:compute_tau}), in the same fashion as we described for the training step of our approach (Algorithm~\ref{alg:disruptiveness}). 
At this stage we distinguish between points of interest that decrease the ranking score (line~\ref{line:increase}) and ones that increase it score (line~\ref{line:decrease}), in order to be able to clearly communicate the effect of certain features to the user. 
In the next step, we calculate the average AP ranking correlation coefficients (lines~\ref{line:avg_tau_upwards} and~\ref{line:avg_tau_downwards}). 
The features that were most disruptive according to this method, are selected as most important features and used as explanations (line~\ref{line:important_values}). 
How many features one reports is again a hyper parameter to tune. 
In our setting LISTEN returns the three most important features. 
As a final step we normalize the labels (line~\ref{line:normalize}), so that we keep the relative importances of each important feature in comparison to the other important features. 
We choose the continuous approach as that allows for a more detailed explanation. 
In our communication to the users we only report the upward pushing labels for reasons explained in Section~\ref{sec:design_listen}.

\begin{algorithm}
\small
	\begin{algorithmic}[1]
		\For{\textbf{each} ranking $\in$ all rankings} \label{line:ranking_loop}
		\For{\textbf{each} item $\in$ ranking}
		\For{\textbf{each} feature $\in$ item}
		\For{\textbf{each} pois value $\in$ all pois values for feature} \label{line:pois_loop}
		\If{pois value $\neq$ feature value} \label{line:value_check}
		\State change feature value
		\State \Call{Calculate$\tau_{AP}$}{new ranking} \label{line:compute_tau}
		\If{new item ranking score $<$ old item ranking score} \label{line:increase}
		\State add $\tau_{AP}$ to upwards pushing $\tau_{AP}$
		\EndIf
		\Else \label{line:decrease}
		\State add $\tau_{AP}$ to downwards pushing $\tau_{AP}$
		\EndIf
		\EndFor
		\State average upwards pushing $\tau_{AP}$ and add to upwards pushing label \label{line:avg_tau_upwards} 
		\State average downwards pushing $\tau_{AP}$ and add to downwards pushing label \label{line:avg_tau_downwards}
		\EndFor
		\State choose most important feature values \label{line:important_values}
		\State normalize labels \label{line:normalize}
		\EndFor
		\EndFor	
	\end{algorithmic}
	\caption{LISTEN explaining phase - Make labels with importance scores per feature value}
	\label{alg:explaining phase}
\end{algorithm}
\vspace*{-10pt}

\subsection{Q-LISTEN to speed up LISTEN}
\label{subsec:q_listen}
The computational complexity of LISTEN is $\mathcal{O}(dnm)$, with $d$ the number of documents in the ranking, $n$ the number of features per document, and $m$ the number of perturbations of feature values per feature.
This is too high to run LISTEN in production in real-time. 
Therefore, we introduce Q-LISTEN. 
For Q-LISTEN we train a multilayer perceptron that learns to generate the explanations produced by LISTEN in a supervised setting. 
During the training, validation and testing phase, we use the ranking data as input and keep the data that we use for each state isolated. 
We use the explanations for these rankings, constructed by LISTEN, as the corresponding labels. 
The ranking data is represented as a matrix of feature values. 
Each row in the matrix represents a new item, ranked in a decreasing order of importance. 
Each column in the matrix represents the value for a given feature. 
We flatten the matrix to provide our network with data it can work with. 
In a real-time environment we only need our trained Q-LISTEN network to produce explanations for incoming new rankings.

In experiments on the data described above (see Section~\ref{sec:experimental_setup} for a more detailed description), Q-LISTEN receives a testing accuracy of $98.7\%$, i.e., Q-LISTEN is very well able to learn the latent explaining function.
It is worth investigating to what extent it improves the speed of the pipeline. 
On a simple notebook (8GB RAM, i5-5200U processor) it takes around 30 seconds to generate listwise reasons for an item, given that the points of interest were already found. 
Using the neural network, this only takes around 1 millisecond. 
Given that we have to generate millions of reasons at production time, this is a speed-up that changes the run time from ``infeasible to run in production'' to ``perfectly fine to run in production.'' 

%% file: 04-experimental-setup.tex

\section{Experimental setup}
\label{sec:experimental_setup}
As explained in Section~\ref{subsection:problem-setting}, we use a production news recommender to answer both of our research questions. In this section we give a description of the heuristic baseline, continue with a description of the data, and present the design of our neural network that we use to speed up LISTEN. At the end of this section we present our experimental setup to answer our second research question.

\subsection{Baseline -- Heuristic reasons}
The production news recommender we employ for our experiments already uses some heuristic justifications as reasons, given in Table~\ref{tab:heuristic_reason_mapping}. We use these as one of the baselines of our research. Properties of the articles and the user are compared and from this comparison reasons are constructed. An example of this could be a long article that is recommended to a user. This particular user may tend to read long articles in general and therefore a justification could be something along the lines of ``\emph{because you seem to like longer articles}.'' Of course, this is not necessarily the real descriptive explanation for why a user sees this article. That means this approach is not faithful. Our new listwise approach solves this issue.

\begin{table}[t]
	\centering
	\caption{Heuristic reasons used in Baseline, numbered.}
    \resizebox{.9\columnwidth}{!}{
	\begin{tabularx}{\columnwidth}{lX}
		\toprule
		\textbf{No.}    & \textbf{Reason description}  \\ 
		\midrule
		0    & Because you often read about CHANNEL. \\ 
		1	& Because often read from PROVIDER. \\
		2	& Because often read from AUTHOR. \\
		3    & Because you are interested in CHANNEL. \\ 
		4	& Because we think CHANNEL could be interesting for you. \\
		5    & Because we think PROVIDER could be interesting for you. \\ 
		6    & Because you follow CHANNEL \\
		7    & Because you follow PROVIDER. \\ 
		8   & Because you seem to like a long read every now and then. \\
		9   & The editors really liked this piece. \\ 
		10    & According to the editors, this is one of the best stories of the day. No matter your preferences. \\
		\bottomrule
	\end{tabularx}
	\label{tab:heuristic_reason_mapping}
    }
    \vspace{-15pt}
\end{table}

\subsection{Data}
We extract around 30Gb of historical feature data of users. Amongst others, this data contains all feature values for all items for approximately 5,500 users. This includes both active and less active users. It is important to have a good mix between these two groups, perhaps even slightly biased towards active users, as non-active users will all have very similar feature values. 

From this data we select $100$ users (active and less active) that we use for our training data in the training step of LISTEN. We use the rest of the data for the explaining step. After the explaining step this part of the data is again divided into training, validation and test data to build the speed up step with the neural networks. 

\subsection{Feedforward neural networks}
We train a straightforward feed forward four layer perceptron with ReLU activations to parameterize our explaining space. The dimensionality of each of the layers is $100$. We use $l2$ weight regularization and a dropout rate of $0.1$. We initialize the weights with the Xavier initializer \citep{glorot2010understanding}. We train our network for $6000$ iterations with a batch size of $50$ and use the Adam optimizer \citep{kingma2014adam} with a learning rate of $2e^{-4}$. We use a standard mean squared error as our loss function.

\subsection{A/B-test to answer RQ2}
Our main goal is to produce faithful explanations that explain a ranking, specifically the ranking of news articles produced by the recommender system of ANONIMYZED. 
Our goal is not to change users' reading behavior by our explanations (e.g., we do not mean to convince users to read a certain article by showing an explanation). 
From the company's perspective, it is very important to avoid any negative reading effects that may occur because of the explanations.
Therefore we investigate whether the reading behavior of the users of the news recommender differs with the explanation system they are exposed to. In order to do so, we run an A/B-test on all users who receive a personalized selection of news articles. Our test consists of two groups:
\begin{enumerate}[nosep,leftmargin=14pt]
\item a group of users that receive the heuristic reasons, 
\item a group of users that receive the reasons produced by Q-LISTEN. 
\end{enumerate}
The users are equally divided over the two groups. We run the A/B-test for fourteen days and by stratified sampling of users we make sure that the users are equally divided over groups in terms of their reading behavior before the start of the test. 

%% file: 05-results.tex

\section{Results}
\label{sec:results}
We look into the faithfulness of our explainer to answer RQ1 and we answer RQ2 by analyzing the results of our A/B-test.\footnote{We originally performed our analysis with three groups. We observed no differences whatsoever between these three groups. We choose to report on only the two most insightful groups in this paper. All numbers are from our original analysis, to ensure the validity of our conclusions.}

\subsection{RQ1 -- Are explanations faithful?}
As a first step, we want to find out whether LISTEN produces faithful explanations. We test this with the ranking function and the ranking that we introduced in Section~\ref{sec:introduction}.

Recall from Section~\ref{sec:introduction} that we want LISTEN to find that for both item $d_0$ and $d_2$ feature $x_1$ was most important. We have two steps that we need to validate. 
First, we need to verify whether the general idea of changing feature values results in the correct behavior. 
Secondly, we need to verify whether using the interest points that we find, also leads to the expected behavior. 

\paragraph{Changing feature values}
First, we test our approach of changing feature values, without selecting points of interest. We change feature values in their entire domain, with steps of $0.01$. We find exactly the behavior we were expecting. The $\tau_{AP}$ values for the ranking are found to be
\vspace{-7pt}
\[
	\begin{array}{l l l l }
	& x_0	& x_1	& x_2 \\
	d_0 & 1.0   & 0.83  & 1.0 \\
	d_1 & 0.64  & 0.67  & 0.99 \\
	d_2 & 1.0   & 0.83  & 1.0
	\end{array} 
\]
Now, looking at the first and the last items in the ranking we see exactly what we predicted in Section~\ref{sec:introduction}. The ranking cannot change by changing feature $x_0$ or $x_2$ in their valid ranges. The ranking can change if we change value $x_1$ and this causes $x_1$ to have the lowest $\tau_{AP}$ value.

\paragraph{Using points of interests}
The first validation step is a sanity check. 
We also need to validate whether our approach of using only certain points of interest is faithful. 
To this end we construct dummy ranking data. 
We use the three features that were introduced above and their domains. We also use the same scoring function. 
We make data points (i.e., ``items'') by randomly sampling feature values for each feature. 
We sample from a range with steps of $0.01$. 

We want to find out whether the number of users in our data and the number of data points per user influence the results. 
Therefore, we make data for multiple numbers of users and a range of data points per user. 
For the number of users we choose from $[5, 10, 20, 100]$ and for the number of data points we choose values from $[5, 10, 20, 40, 60, 80, 100, 120, 150]$. 
We find interest points and compute the most important feature values for our known sample ranking based on these interest points. 
We compute the accuracy, i.e., how often our approach returns the correct feature values. 
Because our approach is not deterministic, as we randomly choose values for the feature values in the data points, we compute the accuracy twenty times per setting (i.e., we construct twenty datasets per setting) and average these.

\begin{figure}[h]
	\centering
	\begin{tikzpicture}
	\begin{axis}[align=center,
	title={ },
	xlabel={Number of data points},
	xlabel near ticks,
	ylabel={Average accuracy},
	ylabel near ticks,	
	legend pos=south east,
	ymajorgrids=true,
	grid style=dashed,
	width=\linewidth,
	ymin=0,
	ymax=1,
	height=4.3cm,
	legend cell align=left,
	]
	
	\addplot[
	color=blue,
	thick,
	]
	coordinates {(5, 0.85)
		(10, 0.8833333333333332)
		(20, 0.85)
		(40, 0.8666666666666666)
		(60, 0.7999999999999999)
		(80, 0.9)
		(100, 0.7833333333333331)
		(120, 0.85)
		(150, 0.8833333333333332)
	};
	\addplot[
	color=red,
	thick,
	]
	coordinates {(5, 0.8833333333333332)
		(10, 0.8499999999999999)
		(20, 0.85)
		(40, 0.8833333333333332)
		(60, 0.8333333333333333)
		(80, 0.8333333333333333)
		(100, 0.8666666666666666)
		(120, 0.8833333333333334)
		(150, 0.8833333333333335)
	};
	\addplot[
	color=green,
	thick,
	]
	coordinates {(5, 0.9333333333333333)
		(10, 0.8666666666666666)
		(20, 0.8333333333333333)
		(40, 0.8833333333333334)
		(60, 0.8333333333333333)
		(80, 0.8333333333333333)
		(100, 0.8333333333333333)
		(120, 0.8333333333333333)
		(150, 0.7666666666666664)
	};
	\addplot[
	color=yellow,
	thick,
	]
	coordinates {(5, 0.9333333333333332)
		(10, 0.8166666666666667)
		(20, 0.7999999999999999)
		(40, 0.9)
		(60, 0.8666666666666666)
		(80, 0.9)
		(100, 0.8666666666666666)
		(120, 0.85)
		(150, 0.8717948717948718)
	};	
	\legend{5 users,
		10 users,
		20 users,
		100 users}	
	\end{axis}
	\end{tikzpicture}
	\caption{Average accuracy when finding points of interest based on different numbers of users and different numbers of data points.}
	\label{fig:average_accuracy_evaluation}
\end{figure}
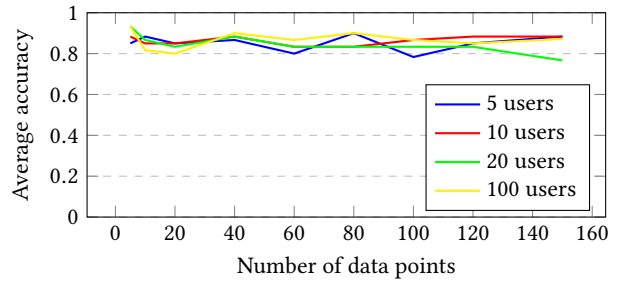

Figure~\ref{fig:average_accuracy_evaluation} shows the accuracy scores per number of users per number of data points. We see no vital differences between different settings. We do see that we do not have a $100\%$ score at all times. These lower scores are most often caused by the reason returned for $d_1$. The disruptiveness of feature $x_0$ and $x_1$ are quite similar for $d_1$ and sometimes $x_0$ is chosen as the most important feature value. 

In our real data, i.e., the historical feature data, the disruptiveness of the different features can also be very similar. This data, however, is a lot more structured than the random data that we have constructed for this dummy experiment. Therefore we can a lot better rely on the found disruptiveness of feature values, as long as we make sure that we use a representative sample of the data. In our case, this means that we need to include active users and less active users. Secondly, if features have very similar disruptive values, these features are of very similar importance in the end. This also reduces the impact on faithfulness reporting the feature value that was actually slightly less important than another feature value, especially as we calculate more than one important feature as reason.  

\subsection{RQ2 -- Is reading behavior impacted?}
\begin{figure}[h]
	\vspace{-5pt}
	\begin{tikzpicture}
	\begin{axis}[align=center,
	legend pos=south west,
	xlabel={Day},
	xlabel near ticks,
	ylabel style={align=center},
	ylabel={Fraction of \\ all reads},
	ylabel near ticks,
	ymin=0.0e-2,
	ymax=3.2e-2,
	ymajorgrids=true,
	grid style=dashed,
	width=\linewidth,
	height=4.3cm,
	legend cell align=left,
	]
	
	\addplot[
	color=blue,
	thick,
	]
	coordinates {(1, 0.02719731653)
		(2, 0.03132224287)
		(3, 0.02577701222)
		(4, 0.02397896741)
		(5, 0.02283063627)
		(6, 0.02559569678)
		(7, 0.02606409501)
		(8, 0.02201471677)
		(9, 0.02603387577)
		(10, 0.02514240817)
		(11, 0.02216581297)
		(12, 0.01850928486)
		(13, 0.02237734766)
		(14, 0.02384298083)
	};
	\addplot[
	color=green,
	thick,
	]
	coordinates {(1, 0.02559569678)
		(2, 0.02966018464)
		(3, 0.02296662285)
		(4, 0.02204493601)
		(5, 0.02163697626)
		(6, 0.02573168336)
		(7, 0.02675913754)
		(8, 0.02107792031)
		(9, 0.02384298083)
		(10, 0.02444736564)
		(11, 0.0207001798)
		(12, 0.01597086865)
		(13, 0.02060952208)
		(14, 0.02425094057)
	};
	\legend{Heuristic,
		Q-LISTEN}
	\end{axis}
	\end{tikzpicture}
	\caption{Reads of users per day during the experiment. Only users with a personalized selection are included. Normalized over all days and all groups.}
	\label{fig:ab_test_after_personalized}
    \vspace{-5pt}
\end{figure}
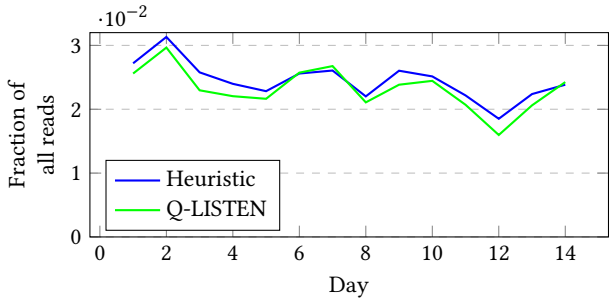

Figure~\ref{fig:ab_test_after_personalized} shows the number of reads per day of users in both groups. The results are normalized for competitiveness reasons. We compute the significance of the differences between the two groups with a randomization test and find that none of the results are significant. 

We also compare the number of reasons users see and how that affects their behavior. For example, we look at how often users open an article within two minutes after seeing a reason. This may indicate that the reason contributed to their decision to open an article. We also look at the number of times users look at a reason within twenty minutes after reading an article. This may indicate that users are wondering why this particular article was selected for them. Table~\ref{tab:reasons_seen} shows these results. Again we computed the significance scores and none of the observed differences were found to be significant. 

\begin{table}[h]
	\centering
	\caption{Reasons seen in both groups and the effects on user behavior.}
    \resizebox{.9\columnwidth}{!}{
	\begin{tabularx}{\columnwidth}{p{4.25cm} r@{.}l r@{.}l}
		\toprule
		& \multicolumn{2}{c}{\textbf{Heuristic}}  & \multicolumn{2}{c}{\textbf{Q-LISTEN}}  \\ 
		\midrule                  
		Percentage of reasons seen (\% of total reads in group)  			& 3&55    & 3&51 	 \\  
		Percentage of reasons seen (\% of total reasons)  					& 34&1    & 32&0 	 \\  
		Reasons per user (of users that see reasons) 						& 1&83 ($\pm$1.61) & 1&72 ($\pm$2.19) \\
		Article opened within two minutes after seeing a reason (\% of all reasons seen) 		& 12&1	& 11&6	 \\ 
		Reasons seen within twenty minutes after opening an article (\% of all reasons seen) 	& 11&0	&  10&7 	 \\ 
		\bottomrule                    
	\end{tabularx}
	\label{tab:reasons_seen}
    }
\end{table}
    
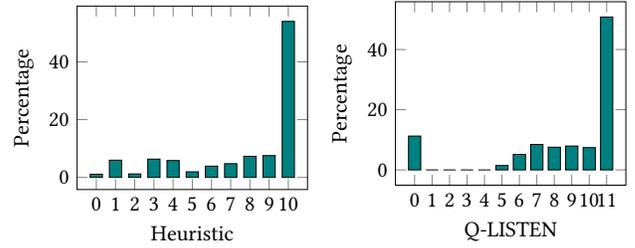
\begin{figure}
\resizebox{.49\linewidth}{!}{
	\begin{tikzpicture}
	\begin{axis}[
	xtick=data,
	bar width=5,
	ylabel=Percentage,
	ylabel near ticks,
	xlabel=Heuristic,
	xlabel near ticks,
	height=120
	]
	\addplot[ybar,fill=teal] coordinates {
		(0, 1.10)
		(1, 5.97)
		(2, 1.22)
		(3, 6.34)
		(4, 5.85)
		(5, 1.95)
		(6, 3.90)
		(7, 4.76)
		(8, 7.31)
		(9, 7.56)
		(10, 54.02)
	};
	\end{axis}
	\end{tikzpicture}
}
\resizebox{.49\linewidth}{!}{
	\begin{tikzpicture}
	\begin{axis}[
	xtick=data,
	bar width=5,
	ylabel=Percentage,
	ylabel near ticks,
	xlabel=Q-LISTEN,
	xlabel near ticks,
	height=120
	]
	\addplot[ybar,fill=teal] coordinates {
		(0, 11.24)
		(1, 0.)
		(2, 0.)
		(3, 0.)
		(4, 0.)
		(5, 1.46)
		(6, 5.16)
		(7, 8.47)
		(8, 7.54)
		(9, 7.93)
		(10, 7.41)
		(11, 50.8)
	};
	\end{axis}
	\end{tikzpicture}	
}
	\caption{Reasons clicked before opening the article, see Table~\ref{tab:features_number_mapping} and~\ref{tab:heuristic_reason_mapping} for a mapping.}
	\label{fig:most_seen_reasons}
    \vspace{-10pt}
\end{figure}

Figure~\ref{fig:most_seen_reasons} shows how often the individual reasons are seen in each group. The numbers of the reasons correspond to the mapping presented in Table~\ref{tab:features_number_mapping} and~\ref{tab:heuristic_reason_mapping}. Reason $9$, $10$ and $11$ in the  Q-LISTEN group respectively correspond to two reasons that are added if users see articles based on a diversification algorithm and a must-read reason that users see for the \textit{must-reads} (reason 10 in Table~\ref{tab:heuristic_reason_mapping}).  We can see that the must-read reason is often shown. Must-reads occur at the beginning of a users personal selection which is likely to explain this peak. 
These results show that the reading behavior of users is not affected by the type of algorithm and that it is safe to use LISTEN, as the most faithful explainer, in production.

%% file: 06-discussion-and-conclusion.tex

\section{Discussion and Conclusion}
\label{sec:disc_and_con}
In this study, we have investigated the explainability of ranking algorithms. To this end, we introduced LISTEN and Q-LISTEN. LISTEN finds the most important features for an item's position in the ranking and returns these as explanations. Q-LISTEN allows us to generate explanations for items in the ranking in production in real time, by using a neural network that is trained to learn the explanation space generated by LISTEN. An A/B-test with reasons produced by different types of explanation systems showed that the reading behavior of users does not differ depending on the type of explanations they see. 
This shows that it is safe to use (Q-)LISTEN in production. (Q-)LISTEN is the only method to produce faithful reasons for the current task. 
Therefore, from a transparency point of view, (Q-)LISTEN outperforms the baseline and is the preferred method to use. 
Although we have tested (Q-)LISTEN in the context of rankings for a news recommender, the approach also generalizes to other ranking and recommender systems.  
Additional research needs to focus on the explainability of systems that make use of less interpretable features, as it will be more difficult to explain these features to users. 
Also more research needs to be conducted on systems that use more features than the current one or features that are correlated as they may interact.
Taking all these features and combinations of these features into account increases the number of comparisons LISTEN needs to make and it blows up the space that is to be learned by the neural network. 
We need to find out how this affects the ability of the network to learn the underlying space.